\documentclass{article}

\PassOptionsToPackage{numbers,compress}{natbib}
\usepackage[preprint]{neurips_2020}

\usepackage[utf8]{inputenc} % allow utf-8 input
\usepackage[T1]{fontenc}    % use 8-bit T1 fonts
\usepackage{hyperref}       % hyperlinks
\usepackage{url}            % simple URL typesetting
\usepackage{booktabs}       % professional-quality tables
\usepackage{amsfonts}       % blackboard math symbols
\usepackage{amsmath}  
\usepackage{amssymb}  
\usepackage{bbding}
\DeclareMathOperator{\E}{\mathbb{E}}

\usepackage{nicefrac}       % compact symbols for 1/2, etc.
\usepackage{microtype}      % microtypography
\usepackage{verbatim}     % for multiple line comments
\usepackage{mathtools}

\usepackage{multirow}
\usepackage{xcolor}
\usepackage{kotex}
\usepackage{booktabs}
\usepackage{amssymb}

\usepackage{booktabs}
\usepackage{caption}
\usepackage{subfig}
\usepackage{tabularx}

\usepackage{varwidth}

\usepackage{textcomp} % for microtype warning

\usepackage{floatrow}
\usepackage{booktabs}

\newfloatcommand{capbtabbox}{table}[][\FBwidth]
\title{README: REpresentation learning by fairness-Aware Disentangling MEthod}

\author{%
  Sungho Park\thanks{qkrtjdgh18@yonsei.ac.kr} \\
  Department of Computer Science\\
  Yonsei University\\
  \texttt{qkrtjdgh18@yonsei.ac.kr} \\
  \And
  Dohyung Kim \\
  Department of Computer Science\\
  Yonsei University\\
  \texttt{dohkim02@yonsei.ac.kr} \\
  \AND
  Sunhee Hwang \\
  Department of Computer Science\\
  Yonsei University\\
  \texttt{sunny16@yonsei.ac.kr} \\
  \And
  Hyeran Byun \\
  Department of Computer Science\\
  Yonsei University\\
  \texttt{hrbyun@yonsei.ac.kr} \\
}

\begin{document}

\maketitle

\begin{abstract} Fair representation learning aims to encode invariant representation with respect to the protected attribute, such as gender or age. In this paper, we design Fairness-aware Disentangling Variational AutoEncoder (FD-VAE) for fair representation learning. This network disentangles latent space into three subspaces with a decorrelation loss that encourages each subspace to contain independent information: 1) target attribute information, 2) protected attribute information, 3) mutual attribute information. After the representation learning, this disentangled representation is leveraged for fairer downstream classification by excluding the subspace with the protected attribute information. We demonstrate the effectiveness of our model through extensive experiments on CelebA and UTK Face datasets. Our method outperforms the previous state-of-the-art method by large margins in terms of equal opportunity and equalized odds.
\end{abstract}

\section{Introduction}
Artificial Intelligence (AI) is one of the most popular research fields involving computer vision, natural language processing, robotics and etc. Although lots of AI models show a great performance nowadays, several studies find out that the models still output discriminatory outcomes for gender or race~\cite{racial_face,tuning,balance}. Previous studies define an attribute, which people should not be discriminated against by, as the protected attribute and propose various fairness methods~\cite{programmer,snowboard,jugde,10.1145/3278721.3278779,Louizos2016TheVF,NIPS2006_3110}. Some studies of them propose fair representation learning methods that encode fair latent variables in terms of the protected attribute and exploit it for various downstream tasks~\cite{10.5555/3294771.3294827,10.1145/3278721.3278779,Fairgan,Louizos2016TheVF,NIPS2006_3110,10.1145/3306618.3314243}. Specifically, ~\cite{pmlr-v97-creager19a} shows disentangled representation learning~\cite{Higgins2017betaVAELB,pmlr-v80-kim18b,NIPS2018_7527}, which separates the latent variables into several subspaces independent of each other, can be utilized for fair representation learning. They disentangle representation into the subspaces: 1) sensitive latents that have high mutual information with the protected attribute labels and 2) non-sensitive latents that are independent of sensitive latents. Then, only non-sensitive latents are leveraged for downstream classification tasks to achieve fair classification results.

\begin{figure*}[t]
  \caption{The motivation of our work. The \textcolor{red}{Red circle} and \textcolor{blue}{blue circle} denote information on the protected attribute and target attribute, respectively. This figure conceptually shows how each latent subspace of previous method~\cite{pmlr-v97-creager19a} (sensitive latents and non-sensitive latents) (a) and our method (PAL, TAL, and MAL) include the information. Sensitive latents contain some target attribute information desired to be included in non-sensitive latents} 
  
  \centering
  \includegraphics[clip=true, width=0.99\textwidth]{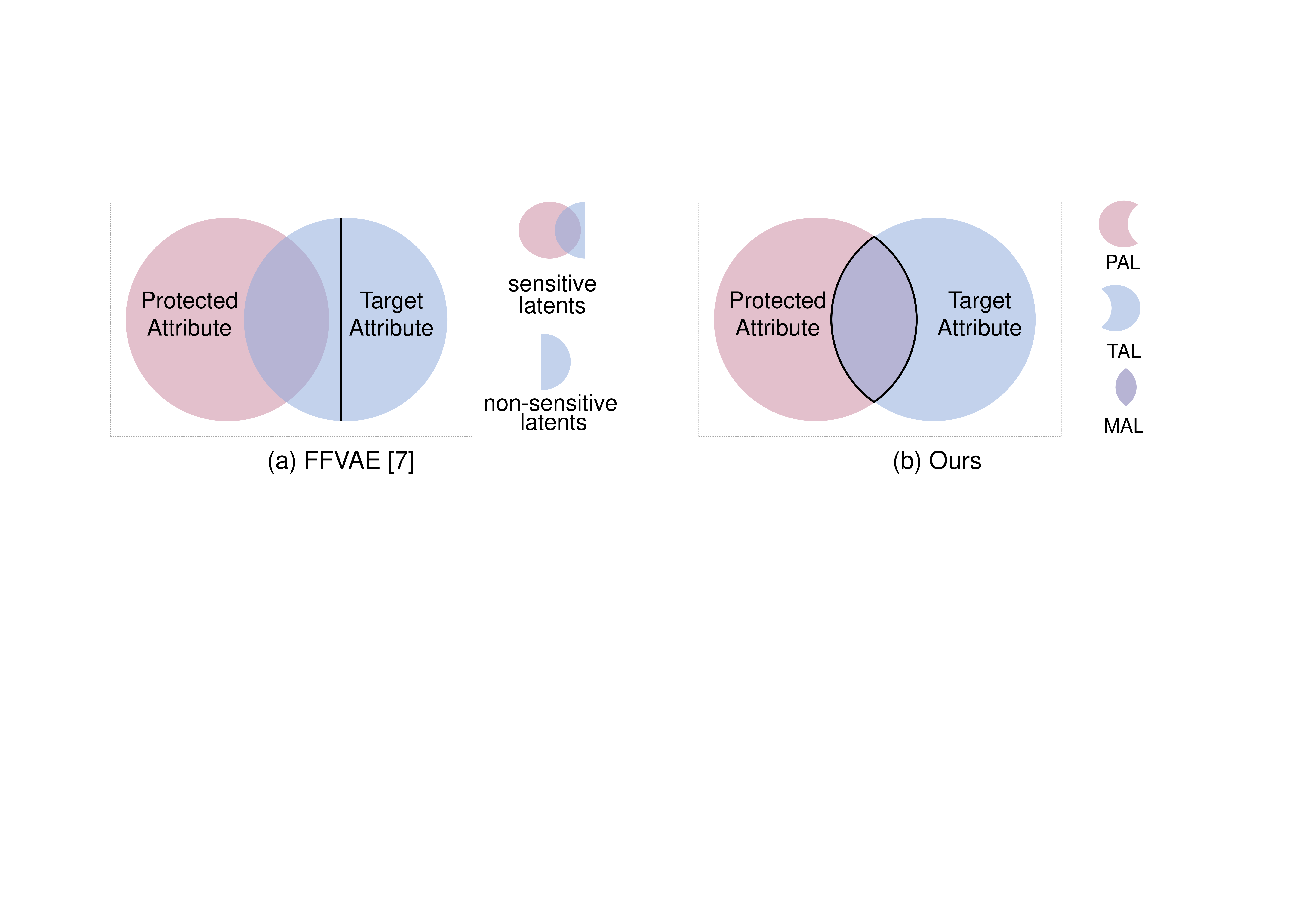}
  \label{fig:intro}
\end{figure*}

However, as pointed out in~\cite{DBLP:journals/corr/LouizosSLWZ15,10.5555/2832747.2832823}, it is challenging to learn non-sensitive latents to have discriminative information for downstream target tasks, since it is learned without the supervision of target labels. In addition, we empirically figure out that sensitive latents contain some target attribute information desired to be included in non-sensitive latents, which is conceptually represented in Figure~\ref{fig:intro}. The experiment is designed to perform downstream classifications using each sensitive and non-sensitive latents. Table~\ref{table:intro_table} shows that classifications using sensitive latents show better performances for both the target and protected attributes than using non-sensitive latents. It indicates that some target attribute information is included in sensitive latents and it is removed since sensitive latents are excluded in downstream tasks.

\begin{table*}[t]
\caption{Comparison between~\cite{pmlr-v97-creager19a} and ours. We perform downstream classification task on CelebA dataset using each subspace of the previous method~\cite{pmlr-v97-creager19a} (left) and ours (right) for the target and protected attributes. The target attribute (TA) and protected attribute (PA) are set to attractive and male, respectively. Sensitive latents show better classification accuracy on both TA and PA than non-sensitive latents. In ours, on the other hand, TAL and PAL shows better performance than each other on TA and PA, respectively.}
\begin{center}
\resizebox{0.80\textwidth}{!}{
\begin{tabular}{ccclccc}
\toprule
\multicolumn{1}{c|}{} & \multicolumn{2}{c}{FFVAE~\cite{pmlr-v97-creager19a}} &  & \multicolumn{3}{c}{Ours} \\ \cline{2-3} \cline{5-7} 
\multicolumn{1}{c|}{} & sensitive latents & non-sensitive latents &  & TAL & \multicolumn{1}{c|}{PAL} & MAL \\ \cline{1-3} \cline{5-7} 
\multicolumn{1}{c|}{TA (Acc.)} & 67.4 & 62.7 &  & 60.3 & \multicolumn{1}{c|}{56.6} & 67.2 \\ \cline{1-3} \cline{5-7} 
\multicolumn{1}{c|}{PA (Acc.)} & 76.7 & 69.1 &  & 63.0 & \multicolumn{1}{c|}{67.3} & 68.9 \\ \bottomrule
\end{tabular}
}
\end{center}
\label{table:intro_table}
\end{table*}

To mitigate the problem aforementioned, we design a Fairness-aware Disentangling Variational AutoEncoder (FD-VAE). Our method disentangles the latent variables into three independent subspaces: the first subspace is related to the target attribute information (TAL), the second one is related to the protected attribute information (PAL), and the last one is related to the mutual attribute information (MAL). In addition, we propose a decorrelation loss to encourage TAL to have only the target attribute information and PAL to have only the protected attribute information as shown in Figure~\ref{fig:intro}b. MAL encourages to decorrelate the information between TAL and PAL by including the mutual attribute information that is predictive to both the attributes. 

When performing downstream classification tasks, we exclude PAL for fair classification with respect to the protected attribute. In addition, we transform MAL into latent variables where the protected attribute information is removed. Then, We utilize TAL and transformed MAL together for a classification, which enables more discriminative classification than using only TAL.
 
In the experiments section, we compare our method to previous disentangled representation learning methods on CelebA and UTK Face datasets~\cite{liu2015faceattributes,zhifei2017cvpr}. To evaluate fairness in downstream classification tasks, we utilize two metrics, equal opportunity and equalized odds~\cite{10.5555/3157382.3157469}. In addition, we propose a \textit{equalized~accuracy} to measure unbiased classification accuracy on a skewed test dataset. On both the datasets, our method shows the best performance in terms of fairness and comparable performances to previous methods in terms of accuracy. It indicates ours shows better trade-off performances between fairness and accuracy than previous models. In addition, we validate the contribution of each component of our method through ablation study on CelebA dataset. In short, the main contributions of our paper are as follows:
\begin{itemize}
    \item We propose a novel FD-VAE to disentangle the latent variables into three independent subspaces related to the target attribute information (TAL), protected attribute information (PAL), and mutual attribute information (MAL).
    \item We propose a loss that decorrelates information of the target attribute and protected attribute in disentangled latent space.
    \item We conduct extensive experiments for downstream classification tasks on CelebA and UTK Face datasets. We exploit equal opportunity and equalized odds to evaluate fairness and suggest \textit{equalized~accuracy} to evaluate unbiased classification accuracy. Our method outperforms the previous state-of-the-art method by large margins in terms of fairness on both datasets.
\end{itemize} 

\section{Related Work}

\subsection{Disentangled Representation Learning}
Many studies~\cite{Higgins2017betaVAELB,pmlr-v80-kim18b,NIPS2018_7527,8099624,8237442,Zhu2014MultiViewPA,image_generation1,styletransfer1,styletransfer2} propose disentangled representation learning methods to learn latent variables independent of each other. ~\cite{Higgins2017betaVAELB} proves KL-divergence term in the VAE objective function encourages latent variables to be disentangled and proposes $\beta$-VAE to weight this term with larger hyperparameter $\beta~(>1)$. However,~\cite{pmlr-v80-kim18b} indicates that $\beta$-VAE has a trade-off between a disentangling performance and reconstruction quality. To reduce the trade-off, they exploit Total Correlation~\cite{10.1147/rd.41.0066}, a measure to estimate the dependency between latent variables, to learn disentangled representation. They approximate it with adversarial learning using discriminator.~\cite{NIPS2018_7527} also optimizes the equivalent objective function to FactorVAE~\cite{pmlr-v80-kim18b} and propose a new stochastic estimation method on Total Correlation, enabling more stable training than FactorVAE.

The concept of disentangled representation is also utilized in various tasks~\cite{8099624,8237442,Zhu2014MultiViewPA,image_generation1,styletransfer1,styletransfer2}. Disentangling methods for a face recognition~\cite{8099624,8237442,Zhu2014MultiViewPA} separate identity representation from other variations like pose or illumination. They exploit the identity representation for a pose-invariant face recognition. In addition,~\cite{image_generation1} proposes generation method for person image and disentangles the input image into three factors (foreground, background, and pose) to manipulate those for image generation.~\cite{translation1,translation2,translation3,translation4} learn style translation models to disentangle representation into style and content representations. In this paper, we leverage disentangled representation for fair representation learning

\subsection{Fair Representation Learning}
Fair representation learning aims to learn fair representation in terms of the protected attribute.~\cite{10.5555/3294771.3294827,10.1145/3278721.3278779,Fairgan} utilize adversarial learning to remove information related to the protected attribute from representation.~\cite{10.5555/3294771.3294827} formulates a representation learning process as an adversarial minimax game, which optimizes the predictor to output target labels and the discriminator to reduce bias to the protected attribute. ~\cite{10.1145/3278721.3278779} propose an adversarial debiasing method to maximize the ability of the predictor for the target class and to minimize the ability of the adversary network for the protected attribute. In addition,~\cite{Fairgan} proposes FairGAN that generates synthetic data unbiased to the protected attribute and trains classifiers using the generated data. The classifiers show fair classification results on the test dataset composed of real data.

Furthermore, several studies propose fair representation learning methods based on the VAE~\cite{Louizos2016TheVF,10.1145/3306618.3314243,pmlr-v97-creager19a}. Variational fair autoencoder~\cite{Louizos2016TheVF} is trained by the penalty term based on Maximum Mean Discrepancy (MMD)~\cite{NIPS2006_3110} and VAE objective function. This method learns representation that is informative on the target label and invariant to the protected attribute.~\cite{10.1145/3306618.3314243} proposes debiasing-VAE (DB-VAE) for a fair face detection. They train the model to identify under-represented data using the latent distribution and adjust the sampling probability of each data to favor the under-represented data. FFVAE~\cite{pmlr-v97-creager19a} is the most similar method to ours. It disentangles representation into sensitive latents and non-sensitive latents. Sensitive latents are learned to include the protected attribute information by the predictiveness term and removed in various downstream classification tasks. In this paper, we find out the limitation that sensitive latents contain some target attribute information with the protected attribute information since the protected attribute is correlated with other attributes in the real world dataset~\cite{biasdataset,emily,jugde}. Compared to FFVAE, our method explicitly decorrelates information of the protected and target attributes, and mitigates this problem.
\begin{figure*}[t]
  \centering
  \includegraphics[clip=true, width=0.99\textwidth]{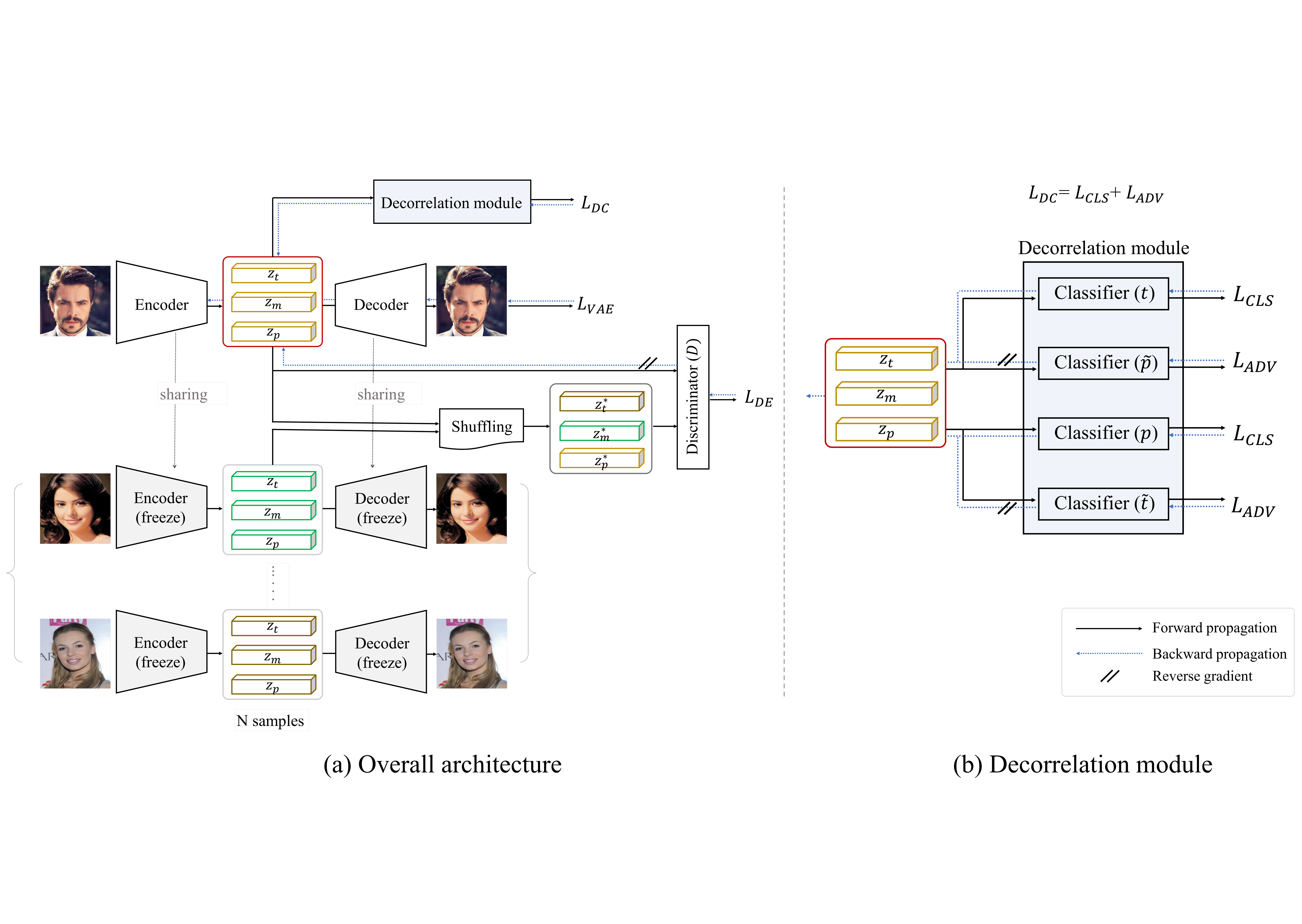}
  \caption{FD-VAE for representation Learning. (a) shows the overall architecture of FD-VAE and (b) shows decorrelation module specifically. PA and TA denote the protected attribute and target attribute, respectively.}
  \label{fig:main_architecture}
\end{figure*}

\section{Proposed Method}
When observed data $X=(x_1,...,x_n)$, target attribute labels $Y_t=(y_{t1},...,y_{tn})$, and protected attribute labels $y_p=(y_{p1},...,y_{pn})$ are given, our goal is to learn latent variables z satisfying three conditions below:
\begin{itemize}
	\item  Maximization of mutual information between $z$ and $X$.
	\item  Disentanglement of $z$ into three subspaces: $z_t$ (TAL), $z_p$ (PAL), and $z_m$ (MAL).
	\item  Decorrelation between the protected and target attribute information in $z$.
\end{itemize}

To achieve the goals, we design FD-VAE that consist of a VAE network, discriminator, and decorrelation module as shown in Figure \ref{fig:main_architecture}. We learn our model by maxmizing the objective function as follows:
\begin{equation}
L_{TOTAL}=L_{VAE}-\alpha L_{DE}-L_{D}-L_{DC},
\end{equation}
where $\alpha$ is a hyperparameter and $L_{VAE}$, $L_{DE}$, $L_{D}$, and $L_{DC}$ denote VAE objective, disentanglement loss, discriminator loss, and decorrelation loss functions, respectively. We specify each loss function in following subsections.

\subsection{VAE}

Our model is based on variational autoencoder (VAE)~\cite{VAE}, which is composed of an encoder and decoder. We learn the VAE network by maximizing the Evidence Lower BOund (ELBO) : 
\begin{equation}
\label{eq:vae}
L_{VAE}= \sum_{i=1}^{n}\E_{q_{\Phi}(z_{t},z_{p},z_{m}|x_i)}[\log{p_\Theta(x_i|z_{t},z_{p},z_{m})}]-KL[q_\Phi(z_{t},z_{p},z_{m}|x_i)||p(z_{t},z_{p},z_{m})],
\end{equation}
where $\Phi$ and $\Theta$ are parameters of the encoder and decoder, respectively. The first term of Equation~\ref{eq:vae} denotes a reconstruction loss that encourages the encoder to map the observed data X into latent variables z and the decoder to reconstruct $X$ from $z$. The latent variables $z$ are sampled from $q_{\Phi}(z|x)=N(\mu_{q_{\Phi}}(x),\sigma_{q_{\phi}}(x)$) using the reparameterization trick, where $\mu$ and $\sigma$ are the outputs of the encoder. The second term indicates a regularization loss that makes the distribution $q_{\Phi}(z|x)$ similar to the gaussian prior distribution $p(z)$ by KL divergence.

\subsection{Disentanglement Loss}

To disentangle the latent variables $z$ into subspaces as $z_{t}$, $z_{p}$, and $z_{m}$, we minimize Total Correlation~\cite{10.1147/rd.41.0066} by following objective function:
\begin{equation}
\label{eq:de}
L_{DE}= KL[q_{\Phi}(z_{t},z_{p},z_{m})||\prod_{j\subseteq S}q_{\Phi}(z_j)]\\=\E_{q_{\Phi}(z_{t},z_{p},z_{m})}[\log\frac{q_{\Phi}(z_{t},z_{p},z_{m})}{\prod\limits_{j\subseteq S}q_{\Phi}(z_j)}],
\end{equation}
where $S=\{t,~p,~m\}$. Total Correlation is one of the popular measures that estimate the dependency between latent variables, and it is demonstrated that minimizing this term encourages latent variables to be disentangled~\cite{pmlr-v80-kim18b,NIPS2018_7527,pmlr-v97-creager19a}. Following the methods in ~\cite{pmlr-v80-kim18b,pmlr-v97-creager19a}, we approximate log density ratio instead of optimizing KL divergence directly by leveraging a discriminator as the following equation:
\begin{equation}
\label{eq:de2}
L_{DE}\approx \E_{q_{\Phi}(z_{t},z_{p},z_{m})}[\log\frac{D(z_t,z_p,z_m)}{1-D(z_t,z_p,z_m)}],
\end{equation}
where $D(z_t,z_p,z_m)$ is the output probability of the discriminator $D$ that classifies the samples from $q_{\Phi}(z_{t},z_{p},z_{m})$ as real and the samples from $\prod\limits_{j\subseteq S}q_{\Phi}(z_j)$ as fake. The encoder is trained for the discriminator not to classify whether if the samples are real or fake. The fake samples $z^{*}$=[$z_t^*$;$z_p^*$;$z_m^*$] are generated by subspace-wise random shuffling within a mini-batch. The following loss is for training the discriminator:
\begin{equation}
 L_D=-\E_{q_{\Phi}(z_{t},z_{p},z_{m})}[\log D(z_t,z_p,z_m)+\log (1-D(z_t^*,z_p^*,z_m^*)].
\end{equation}

\subsection{Decorrelation Loss}
On top of the disentanglement loss that encourages the subspaces to be independent of each other, we introduce a decorrelation loss that specifies which attribute information to be included to each subspace. The decorrelation loss is composed of $L_{CLS}$ and $L_{ADV}$ as follows:
\begin{equation}
L_{DC}=\beta L_{CLS}+\gamma L_{ADV},
\end{equation}
\begin{equation}
L_{CLS}= -\sum_{i=1}^{n}\E_{q_{\Phi}(z_{t},z_{p},z_{m}|x_i)}[\log p_{t}(y_t|z_t))+\log p_{p}(y_p|z_p))], \\
\end{equation}
\begin{equation}
L_{ADV}= \underset{\tilde{t},\tilde{p}}{\max}~\underset{\Phi}{\min}\sum_{i=1}^{n}\E_{q_{\Phi}(z_{t},z_{p},z_{m}|x_i)}[\log p_{\tilde{p}}(y_p|z_t))+\log p_{\tilde{t}}(y_t|z_p))],
\end{equation}
where $t$ and $\tilde{t}$ denote classifiers for the target attribute, and $p$ and $\tilde{p}$ denote classifiers for the protected attribute. $\beta$ and $\gamma$ are hyperparameters. $L_{CLS}$ encourages $z_{t}$ and $z_{p}$ to contain the information of the target attribute and protected attribute, respectively. Meanwhile, $L_{ADV}$ encourages $z_t$ and $z_p$ to exclude information of the protected attribute and target attribute, respectively. However, there is some information that is predictive to both the protected and target attributes. Since the mutual attribute information increases $L_{ADV}$ whether it is included in $z_t$ or $z_p$, we introduce $Z_m$ to include this information. There is no additional mapping function on this subspace $Z_m$, but $L_{DE}$ works indirectly.
\begin{figure*}[t]
  \centering
  \includegraphics[clip=true, width=0.99\textwidth]{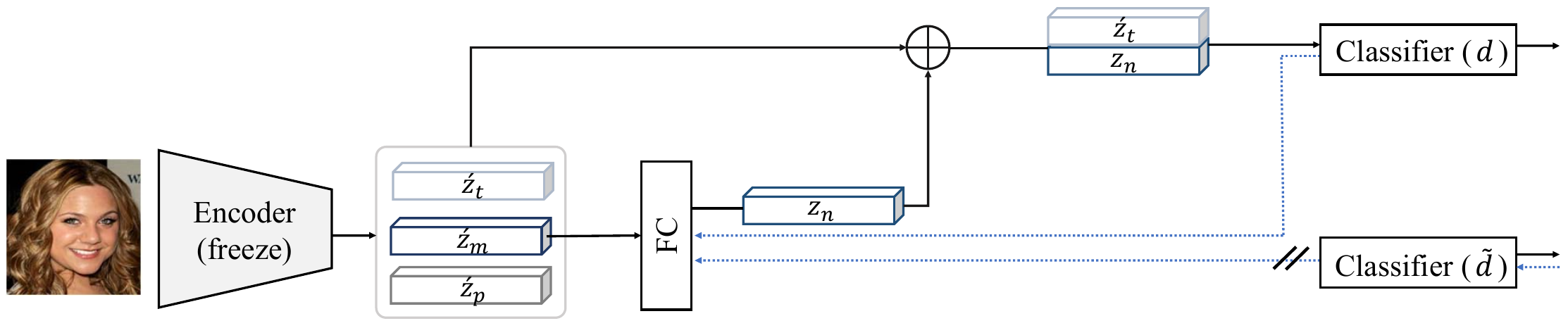}
  \caption{A model for downstream classification tasks. FC denotes a singe fully connected layer and $\oplus$ is element-wise summation.}
  \label{fig:main_architecture2}
\end{figure*}

\subsection{Downstream Classification Network}
After learning a fair representation, we perform downstream classification tasks to predict the target attribute. The overall architecture of our classification model is shown in Figure~\ref{fig:main_architecture2}. The loss function to optimize the classification model is defined by:
\begin{equation}
 \label{equ:DCLS}
L_{DCLS}= \underset{d,f}{\max}~ \underset{\tilde{d}}{\min}\sum_{i=1}^{n}\E_{q_{\Phi}(\acute{z_{t}},\acute{z_{p}},\acute{z_{m}}|x_i)}[\log p_{d}(y_t|\acute{z_t}\oplus f(\acute{z_m}))-\log p_{\tilde{d}}(y_p|f(\acute{z_m}))],
\end{equation}
where $d$, $\tilde{d}$, and $\oplus$ indicate a target attribute classifier, protected attribute classifier, and element-wise summation. The values of the learned latent variables $\acute{z_t}$, $\acute{z_p}$, and $\acute{z_m}$ are fixed in downstream classification tasks. Firstly, we exclude $\acute{z_p}$ that the protected attribute information is not exploited  for downstream classifications. Then, $\acute{z_m}$ is transformed to latent variables $z_n$ that is irrelevant to the protected attribute by a single fully connected layer $f$. $f$ and $\tilde{d}$ are learned adversarially for this transformation. Finally, $z_n$ and $\acute{z_t}$ are element-wise summed and feed into $d$ for a target classification.

\section{Experiments}

\subsection{Dataset}
We validate our contributions on CelebA and UTK datasets. CelebA dataset is consists of about 200k face images with 40 binary attribute annotations and divided into train, validation, and test sets. We set the protected attributes to male and young, and the target attributes to attractive, wavy hair, and big nose. Specifically, we compose four pairs of attributes considering the correlation between the protected and target attributes: [male, attractive], [male, wavy hair], [young, attractive], and [young, big nose]. 

UTK Face dataset is a face dataset with long age span involving annotations of age, ethnicity, and gender. We set the protected attribute to gender and reclassify the ethnicity annotations into Caucasians and the others, and the age annotations into young and the others(>35). In addition, we divide the dataset into train (10k), validataion (2.4k), and test sets (2.4k). In detail, we compose the train set to have a correlation between the protected and target attributes for fairness study. In respect to the ethnicity, one-fifth of Caucasians are female and four-fifths are male, and the others have the opposite ratio. Likewise, in respect to the age, one-fifth of young are male and four-fifths are female, and the others have the opposite ratio. In addition, we set both the datasets for validation and test to be balanced sets (cf. Appendix A).

\subsection{Evaluation Metrics}
As metrics for fairness, we use equal opportunity and equalized odds which are defined as $|\text{TPR}_{p_0}-\text{TPR}_{p_1}|$ and $\frac{1}{2}[|\text{TPR}_{p_0}-\text{TPR}_{p_1}|+|\text{TNR}_{p_0}-\text{TNR}_{p_1}|]$, respectively, where $p_1$ are data with the positive protected attribute and $p_0$ are vice versa. TPR and TNR are the abbreviation of true positive rate and true negative rate, respectively.

In addition, we suggest \textit{equalized~accuracy} to evaluate unbiased classification accuracy. If train and test set have similar data distributions, the standard accuracy metric benefits models biased to the distribution of train set. On UTK Face dataset,  we solve this problem by comprising a balanced test set, but CelebA dataset provides a fixed test set. Alternatively, we introduce \textit{equalized~accuracy} which has the same effect as the balanced set. This metric is defined as: \text{$\frac{1}{4}[\text{TPR}_{p_0}+\text{TNR}_{p_0}+\text{TPR}_{p_1}+\text{TNR}_{p_1}]$}. More details on \textit{equalized~accuracy} are described in Appendix B.

\subsection{Implementation Details}
The detailed structures of our networks are specified in Appendix C. In our model, TAL, PAL, and MAL are set to 20 dimensions, respectively. In addition, for fair comparison, the representation of all models are set to 60 dimensions. The dimension of sensitive latents and non-sensitive latents in FFVAE are divided into 30. For downstream classification tasks, we remove one or two latents the most correlated with the protected attribute in FactorVAE and $beta$-VAE as in~\cite{pmlr-v97-creager19a}. The hyperparameters $\alpha$, $\beta$, and $\gamma$ are fixed to 50,5 and 10, respectively.

\begin{table*}[t]

\caption{Classification results on CelebA dataset. TA and PA are the abbreviations of the target attribute and protected attribute. We utilize four metrics: equal opportunity (Opp.), equalized odds (Odds), accuracy (ACC.), and \textit{equalized~accuracy} (EAcc.). M,Y,A,W and B denote male, young, attractive, wavy hair, and big nose attributes, respectively.
}
\begin{center}
\resizebox{1.0\textwidth}{!}{
\begin{tabular}{ccccccclcccccc}
\toprule
\multirow{2}{*}{Method} & \multirow{2}{*}{TA} & \multicolumn{2}{c}{PA} & \multirow{2}{*}{Opp. $\downarrow$} & \multirow{2}{*}{Odds $\downarrow$} & Acc. $\uparrow$ &  & \multirow{2}{*}{TA} & \multicolumn{2}{c}{PA} & \multirow{2}{*}{Opp. $\downarrow$} & \multirow{2}{*}{Odds $\downarrow$} & Acc. $\uparrow$ \\ \cline{3-4}  \cline{10-11}  
 &  & M=1 & M=0 &  &  & EAcc. $\uparrow$ &  &  & Y=1 & Y=0 &  &  & EAcc. $\uparrow$ \\ \cline{1-7} \cline{9-14} 
\multirow{2}{*}{VAE~\cite{VAE}} & A=1 & 54.3 & 81.5 & \multirow{2}{*}{27.2} & \multirow{2}{*}{28.7} & 70.6 &  & A=1 & 77.4 & 67.6 & \multirow{2}{*}{9.7} & \multirow{2}{*}{11.3} & 70.9 \\   
 & A=0 & 77.8 & 47.5 &  &  & 65.3 &  & A=0 & 60.8 & 72.7 &  &  & 69.6 \\ \cline{1-7} \cline{9-14} 
\multirow{2}{*}{$\beta-$VAE~\cite{Higgins2017betaVAELB}} & A=1 & 57.7 & 78.8 & \multirow{2}{*}{21.1} & \multirow{2}{*}{22.1} & 68.7 &  & A=1 & 72.9 & 65.6 & \multirow{2}{*}{7.2} & \multirow{2}{*}{5.8} & 67.5 \\   
 & A=0 & 73.1 & 48.8 &  &  & 64.6 &  & A=0 & 61.1 & 65.4 &  &  & 66.4 \\ \cline{1-7} \cline{9-14} 
 \multirow{2}{*}{FactorVAE~\cite{pmlr-v80-kim18b}} & A=1 & 59.7 & 80.4 & \multirow{2}{*}{20.7} & \multirow{2}{*}{23.4} & 69.0 &  & A=1 & 76.0 & 66.8 & \multirow{2}{*}{9.2} & \multirow{2}{*}{9.1} & 68.9 \\   
 & A=0 & 72.8 & 46.6 &  &  & 64.9 &  & A=0 & 58.9 & 68.1 &  &  & 67.5 \\ \cline{1-7} \cline{9-14} 
\multirow{2}{*}{FFVAE~\cite{pmlr-v97-creager19a}} & A=1 & 54.9 & 73.4 & \multirow{2}{*}{17.7} & \multirow{2}{*}{17.5} & 62.7 &  & A=1 & 70.4 & 62.3 & \multirow{2}{*}{8.0} & \multirow{2}{*}{8.2} & 64.2 \\   
 & A=0 & 63.3 & 46.5 &  &  & 59.5 &  & A=0 & 55.3 & 63.6 &  &  & 62.9 \\ \cline{1-7} \cline{9-14}
\multirow{2}{*}{Ours} & A=1 & 66.2 & 67.6 & \multirow{2}{*}{\textbf{1.3}} & \multirow{2}{*}{\textbf{4.9}} & 64.1 &  & A=1 & 76.5 & 72.7 & \multirow{2}{*}{\textbf{3.7}} & \multirow{2}{*}{\textbf{1.9}} & 65.5 \\   
 & A=0 & 64.4 & 55.8 &  &  & 63.5 &  & A=0 & 59.4 & 55.2 &  &  & 66.0 \\ \cline{1-7} \cline{9-14} 
\multicolumn{1}{l}{} & \multicolumn{1}{l}{} & \multicolumn{1}{l}{} & \multicolumn{1}{l}{} & \multicolumn{1}{l}{} & \multicolumn{1}{l}{} & \multicolumn{1}{l}{} &  & \multicolumn{1}{l}{} & \multicolumn{1}{l}{} & \multicolumn{1}{l}{} & \multicolumn{1}{l}{} & \multicolumn{1}{l}{} & \multicolumn{1}{l}{} \\ \cline{1-7} \cline{9-14} 
\multirow{2}{*}{VAE~\cite{VAE}} & W=1 & 5.5 & 52.7 & \multirow{2}{*}{47.1} & \multirow{2}{*}{33.4} & 73.0 &  & B=1 & 58.5 & 75.5 & \multirow{2}{*}{17.0} & \multirow{2}{*}{19.2} & 67.2 \\   
 & W=0 & 98.7 & 79.0 &  &  & 59.0 &  & B=0 & 71.6 & 50.2 &  &  & 64.0 \\ \cline{1-7} \cline{9-14} 
\multirow{2}{*}{$\beta-$VAE~\cite{Higgins2017betaVAELB}} & W=1 & 6.6 & 42.1 & \multirow{2}{*}{35.4} & \multirow{2}{*}{15.3} & 70.7 &  & B=1 & 55.0 & 67.4 & \multirow{2}{*}{12.3} & \multirow{2}{*}{13.9} & 65.0 \\   
 & W=0 & 98.0 & 82.6 &  &  & 57.3 &  & B=0 & 69.2 & 53.6 &  &  & 61.3 \\ \cline{1-7} \cline{9-14}  
\multirow{2}{*}{FactorVAE~\cite{pmlr-v80-kim18b}} & W=1 & 8.1 & 47.5 & \multirow{2}{*}{38.8} & \multirow{2}{*}{28.8} & 71.0 &  & B=1 & 56.0 & 69.2 & \multirow{2}{*}{13.2} & \multirow{2}{*}{14.4} & 64.8 \\   
 & W=0 & 97.6 & 78.7 &  &  & 58.0 &  & B=0 & 68.6 & 53.0 &  &  & 61.7 \\ \cline{1-7} \cline{9-14} 
\multirow{2}{*}{FFVAE~\cite{pmlr-v97-creager19a}} & W=1 & 4.5 & 25.7 & \multirow{2}{*}{21.1} & \multirow{2}{*}{15.4} & 67.7 &  & B=1 & \multicolumn{1}{c}{50.8} & \multicolumn{1}{c}{56.2} & \multirow{2}{*}{5.3} & \multirow{2}{*}{8.5} & 63.0 \\   
 & W=0 & 98.4 & 88.7 &  &  & 54.3 &  & B=0 & \multicolumn{1}{c}{67.3} & \multicolumn{1}{c}{58.7} &  &  & 58.2 \\ \cline{1-7} \cline{9-14} 
\multirow{2}{*}{Ours} & W=1 & 3.8 & 17.2 & \multirow{2}{*}{\textbf{13.4}} & \multirow{2}{*}{\textbf{8.9}} & 66.1 &  & B=1 & \multicolumn{1}{c}{48.2} & \multicolumn{1}{c}{47.5} & \multicolumn{1}{c}{\multirow{2}{*}{\textbf{1.2}}} & \multicolumn{1}{c}{\multirow{2}{*}{\textbf{1.5}}} & \multicolumn{1}{c}{63.9} \\   
 & W=0 & 97.5 & 93.1 &  &  & 52.9 &  & B=0 & \multicolumn{1}{c}{69.6} & \multicolumn{1}{c}{62.6} & \multicolumn{1}{c}{} & \multicolumn{1}{c}{} & \multicolumn{1}{c}{57.7} \\ \bottomrule

\end{tabular}
}
\end{center}

\label{table:main}
\end{table*}

\begin{table*}[t]

\caption{Classification results on UTK Face dataset. G,E, and A indicate gender (1:male, 0:female), ethnicity (1:Caucasian, 0:others), and age (1:young, 0:old), respectively.
}
\begin{center}
\resizebox{0.99\textwidth}{!}{
\begin{tabular}{ccccccclcccccc}
\toprule
\multirow{2}{*}{Method} & \multirow{2}{*}{TA} & \multicolumn{2}{c}{PA} & \multirow{2}{*}{Opp. $\downarrow$} & \multirow{2}{*}{Odds $\downarrow$} & \multirow{2}{*}{Acc. $\uparrow$} &  & \multirow{2}{*}{TA} & \multicolumn{2}{c}{PA} & \multirow{2}{*}{Opp. $\downarrow$} & \multirow{2}{*}{Odds $\downarrow$} & \multirow{2}{*}{Acc. $\uparrow$} \\ \cline{3-4} \cline{10-11}
 &  & G=1 & G=0 &  &  &  &  &  & G=1 & G=0 &  &  &  \\ \cline{1-7} \cline{9-14} 
\multirow{2}{*}{VAE~\cite{VAE}} & E=1 & 70.5 & 54.3 & \multirow{2}{*}{16.1} & \multirow{2}{*}{17.4} & \multirow{2}{*}{65.2} &  & A=1 & 50.8 & 75.5 & \multirow{2}{*}{24.6} & \multirow{2}{*}{29.8} & \multirow{2}{*}{60.7} \\
 & E=0 & 58.6 & 77.4 &  &  &  &  & A=0 & 75.8 & 40.7 &  &  &  \\ \cline{1-7} \cline{9-14} 
\multicolumn{1}{c}{\multirow{2}{*}{$\beta-$VAE~\cite{Higgins2017betaVAELB}}} & E=1 & \multicolumn{1}{c}{72.1} & \multicolumn{1}{c}{59.4} & \multirow{2}{*}{12.7} & \multirow{2}{*}{12.1} & \multirow{2}{*}{60.1} &  & A=1 & \multicolumn{1}{c}{45.3} & 62.5 & \multirow{2}{*}{17.1} & \multirow{2}{*}{20.8} & \multirow{2}{*}{54.3} \\
\multicolumn{1}{c}{} & E=0 & \multicolumn{1}{c}{48.7} & \multicolumn{1}{c}{60.2} &  &  &  &  & A=0 & \multicolumn{1}{c}{67.1} & 42.5 &  &  &  \\ \cline{1-7} \cline{9-14} 
\multirow{2}{*}{FactorVAE~\cite{pmlr-v80-kim18b}} & E=1 & \multicolumn{1}{c}{72.5} & \multicolumn{1}{c}{60.3} & \multirow{2}{*}{12.1} & \multirow{2}{*}{12.9} & \multirow{2}{*}{59.4} &  & A=1 & 43.5 & 66.7 & \multirow{2}{*}{23.2} & \multirow{2}{*}{27.1} & \multirow{2}{*}{54.6} \\
 & E=0 & \multicolumn{1}{c}{45.8} & \multicolumn{1}{c}{59.5} &  &  &  &  & A=0 & 69.6 & 38.5 &  &  &  \\ \cline{1-7} \cline{9-14} 
\multirow{2}{*}{FFVAE~\cite{pmlr-v97-creager19a}} & E=1 & 65.1 & 55.2 & \multirow{2}{*}{9.8} & \multirow{2}{*}{9.1} & \multirow{2}{*}{59.7} &  & A=1 & 45.2 & 58.9 & \multirow{2}{*}{13.6} & \multirow{2}{*}{17.2} & \multirow{2}{*}{54.5} \\
 & E=0 & 54.8 & 64.0 &  &  &  &  & A=0 & 67.4 & 46.5 &  &  &  \\ \cline{1-7} \cline{9-14} 
\multirow{2}{*}{Ours} & E=1 &65.8 & 63.5 & \multirow{2}{*}{\textbf{2.3}} & \multirow{2}{*}{\textbf{1.1}} & \multirow{2}{*}{60.3} &  & A=1 & 47.9 & 45.8 & \multirow{2}{*}{\textbf{2.0}} & \multirow{2}{*}{\textbf{2.9}} & \multirow{2}{*}{54.1} \\
 & E=0 & 56.0 & 56.0 &  &  &  &  & A=0 & 63.3 & 59.4 &  &  &  \\\bottomrule

\end{tabular}
}
\end{center}
\label{table:utk_table}
\end{table*}

\begin{table*}[t]
\caption{Ablation Study on CelebA dataset. We set FFVAE~\cite{pmlr-v97-creager19a} to the baseline (first row). $z_m$ denotes MAL in representation learning and $\acute{z_t}$,$\acute{z_m}$, and $z_n$ denote learned TAL, MAL, and transformed MAL in the downstream classification, respectively. We set the protected attribute to Male (M) and the target attribute to Attractive (A) in this experiment.}
\label{table:ablation}
\begin{center}
\resizebox{0.99\textwidth}{!}{
\begin{tabular}{cccccccccccccc}
\toprule
\multicolumn{3}{c}{Representation learning} &  & \multicolumn{3}{c}{Downstream Classification} &  & \multirow{2}{*}{TA} & \multicolumn{2}{c}{PA} & \multirow{2}{*}{Opp. $\downarrow$} & \multirow{2}{*}{Odds $\downarrow$} & \multirow{2}{*}{Acc. $\uparrow$} \\ \cline{5-7} \cline{1-3} \cline{10-11}
$L_{CLS}$ & $L_{ADV}$ & $z_m$ &  & $\acute{z_t}$ & $\acute{z_m}$ & $z_n$ &  &  & M=1 & M=0 &  &  &  \\ \cline{1-3} \cline{5-7} \cline{9-14} 
\multirow{2}{*}{} & \multirow{2}{*}{} & \multirow{2}{*}{} &  & \multirow{2}{*}{\checkmark} & \multirow{2}{*}{} & \multirow{2}{*}{} &  & A=1 & 54.9 & 73.4 & \multirow{2}{*}{17.7} & \multirow{2}{*}{17.5} & \multirow{2}{*}{62.7} \\
 &  &  &  &  &  &  &  & A=0 & 63.3 & 46.5 &  &  &  \\ \cline{1-3} \cline{5-7} \cline{9-14} 
\multirow{2}{*}{\checkmark} & \multirow{2}{*}{} & \multirow{2}{*}{} &  & \multirow{2}{*}{\checkmark} & \multirow{2}{*}{} & \multirow{2}{*}{} &  & A=1 & 60.7 & 76.4 & \multirow{2}{*}{15.6} & \multirow{2}{*}{16.2} & \multirow{2}{*}{65.3} \\
 &  &  &  &  &  &  &  & A=0 & 64.2 & 47.3 &  &  &  \\ \cline{1-3} \cline{5-7} \cline{9-14} 
\multirow{2}{*}{\checkmark} & \multirow{2}{*}{\checkmark} & \multirow{2}{*}{} &  & \multirow{2}{*}{\checkmark} & \multirow{2}{*}{} & \multirow{2}{*}{} &  & A=1 & 75.7 & 72.3 & \multirow{2}{*}{3.4} & \multirow{2}{*}{5.7} & \multirow{2}{*}{64.1} \\
 &  &  &  &  &  &  &  & A=0 & 58.7 & 50.8 &  &  &  \\ \cline{1-3} \cline{5-7} \cline{9-14} 
\multirow{2}{*}{\checkmark} & \multirow{2}{*}{\checkmark} & \multirow{2}{*}{\checkmark} &  & \multirow{2}{*}{\checkmark} & \multirow{2}{*}{} & \multirow{2}{*}{} &  & A=1 & 68.1 & 66.3 & \multirow{2}{*}{1.8} & \multirow{2}{*}{\textbf{2.1}} & \multirow{2}{*}{60.3} \\
 &  &  &  &  &  &  &  & A=0 & 55.1 & 52.6 &  &  &  \\ \cline{1-3} \cline{5-7} \cline{9-14} 
\multirow{2}{*}{\checkmark} & \multirow{2}{*}{\checkmark} & \multirow{2}{*}{\checkmark} &  & \multirow{2}{*}{\checkmark} & \multirow{2}{*}{\checkmark} & \multirow{2}{*}{} &  & A=1 & 57.9 & 77.7 & \multirow{2}{*}{19.8} & \multirow{2}{*}{20.8} & \multirow{2}{*}{67.1} \\
 &  &  &  &  &  &  &  & A=0 &69.8  & 47.8 &  &  &  \\ \cline{1-3} \cline{5-7} \cline{9-14} 
\multirow{2}{*}{\checkmark} & \multirow{2}{*}{\checkmark} & \multirow{2}{*}{\checkmark} &  & \multirow{2}{*}{\checkmark} & \multirow{2}{*}{} & \multirow{2}{*}{\checkmark} &  & A=1 & 66.2 & 67.6 & \multirow{2}{*}{\textbf{1.3}} & \multirow{2}{*}{4.9} & \multirow{2}{*}{64.1} \\
 &  &  &  &  &  &  &  & A=0 & 64.4 & 55.8 &  &  & \\
 \bottomrule
\end{tabular}
}
\end{center}
\end{table*}

\subsection{Evaluation}
To validate our method, we compare ours with previous methods~\cite{Higgins2017betaVAELB,pmlr-v80-kim18b,pmlr-v97-creager19a}. Table~\ref{table:main} shows the classification results on CelebA dataset. VAE (baseline), which does not consider the disentanglement of latent variables, shows the unfairest performance in all the experiments. The two disentangling methods $\beta$-VAE~\cite{Higgins2017betaVAELB} and FactorVAE~\cite{pmlr-v80-kim18b} improve fairness of results than the baseline. However, it is not significant since they do not leverage the protected attribute in the disentanglement process. FFVAE~\cite{pmlr-v97-creager19a}, which is the state-of-the-art method, achieves better results in terms of both equal opportunity and equalized odds than the other previous methods. Our method shows the fairest performance in all the experiments and outperforms FFVAE by large margins of 16.4$\%$, 7.7$\%$, 4.3$\%$, and 3.1$\%$ at equal opportunity and 12.6$\%$, 6.5$\%$, 6.3$\%$, and 7.0$\%$ at equalized odds. In terms of accuracy and \textit{equalized~accuracy}, ours shows similar performances to FFVAE, and it indicates ours has better trade-off performances between fairness scores and classification accuracy.

The results on UTK Face dataset are shown in Table~\ref{table:utk_table}. Similar to the above, VAE (baseline) shows highly unfair results in terms of both equal opportunity and equalized odds, and the previous methods~\cite{Higgins2017betaVAELB,pmlr-v80-kim18b,pmlr-v97-creager19a} improve fairness scores over the baseline. Our method significantly surpasses the previous methods, showing 2.3$\%$ and $1.1\%$ at equal opportunity, and 2.0$\%$ and $2.9\%$ at equalized odds. 

In conclusion, the experiments on the two datasets indicate that our method decorrelates information of the protected attribute and target attribute better than the others, which causes superior trade-off performance between fairness scores and classification accuracy.

\subsection{Ablation Study}
We conduct ablation study on CelebA dataset to validate the contribution of each component of our method. We set FFVAE~\cite{pmlr-v97-creager19a} to a baseline in this experiment and added each component step by step. The results are shown in Table~\ref{table:ablation}. Firstly, we add $L_{CLS}$ term of the decorrelation loss, which trains TAL by the supervision with the target attribute labels. As a result, classification accuracy and fairness scores are improved than the baseline model. Secondly, we add the total decorrelation loss ($L_{CLS}$+$L_{ADV}$). The information between the target and protected attributes is more clearly decorrelated than using $L_{CLS}$ term only. It significantly improves fairness scores by 12.2\% and 10.5\% in equal opportunity and equalized odds, respectively. 

Then, we experiment three models added with MAL ($z_m$) in representation learning. The first model utilizes only TAL ($\acute{z_t}$) in the downstream classification task. In addition, the second and final models use MAL ($\acute{z_m}$) or transformed MAL ($z_n$) with TAL, respectively. The first model demonstrates that MAL helps to exclude the protected attribute information from TAL, showing the lowest fairness scores. However, it degrades classification accuracy since the mutual attribute information is not utilized for classification. The second and final models show that the transformation of $\acute{z_m}$ is effective. Although utilizing learned MAL without the transformation improves classification accuracy, it highly degrades fairness scores. The final model (FD-VAE) improves classification accuracy while maintaining outstanding fairness scores.

\section{Conclusion}
In this paper, we introduced the Fairness-aware Disentangling Variational Auto Encoder (FD-VAE) that disentangles latent variables into three subspaces including information of the target attribute, protected attribute, and mutual attribute, respectively. By excluding the subspace with protected attribute information, we performed fair downstream classification tasks. In all the experiments on CelebA and UTK datasets, our method achieved the fairest results in terms of equal opportunity and equalized odds.

\section*{Broader Impact}

This work mitigates ethical-social problems caused by Artificial Intelligence (AI) models. Previous AI models have caused social discrimination, such as racism or sexism. For example, Google Photos, one of the visual recognition applications, recognized a couple of African-Americans as a gorilla~\cite{gorilla}. Moreover, Compas algorithm, which is used to predict recidivism by courts in the United States, has judged unfairly on African-American~\cite{propublica}. Our method separates information causing such the discrimination from data representation and makes a fair decision. In addition, our fair representation can be applied to various tasks that the fairness of the results is required. Therefore, our study has a positive impact that mitigates the problems of prejudice in AI models and enables individuals to have fair and plentiful benefits.

\begin{ack}
TBD.
\end{ack}

{
\bibliography{neurips}
\bibliographystyle{abbrvnat}
}

\end{document}

\maketitle

\section{Appendix A: Composition of UTK Face dataset in our experiments}

\begin{table*}[h]
\caption{Our composition of train, validation, and test sets of UTK Face dataset. The figures denote the number of data and G, E, and A denote gender, ethnicity, and age attributes, respectively. }
\label{table:appendixA}
\begin{center}
\resizebox{0.55\textwidth}{!}{
\begin{tabular}{ccclccc}
\toprule
\multicolumn{3}{c}{\textbf{Train set}} & \multicolumn{1}{l}{} & \multicolumn{3}{c}{\textbf{Validation/ Test set}} \\
\multirow{2}{*}{TA} & \multicolumn{2}{c}{PA} & \multicolumn{1}{l}{} & \multirow{2}{*}{TA} & \multicolumn{2}{c}{PA} \\ \cline{2-3} \cline{6-7} 
 & G=1 & G=0 & \multicolumn{1}{l}{} &  & G=1 & G=0 \\ \cline{1-3} \cline{5-7} 
E=1 & 4,000 & 1,000 & \multicolumn{1}{l}{} & E=1 & 600 & 600 \\
E=0 & 1,000 & 4,000 & \multicolumn{1}{l}{} & E=0 & 600 & 600 \\
\cline{1-3} \cline{5-7} 
A=1 & 1,000 & 4,000 & \multicolumn{1}{l}{} & A=1 & 600 & 600 \\
A=0 & 4,000 & 1,000 & \multicolumn{1}{l}{} & A=0 & 600 & 600\\
\bottomrule
\end{tabular}
}
\end{center}
\end{table*}
In our experiments, we compose train, validation, and test sets of UTK Face dataset~\cite{zhifei2017cvpr} as shown in Table~\ref{table:appendixA}. Firstly, we compose a skewed dataset since it is not challenging to perform fair classification when representation is learned on a balanced dataset. For example, we set more Caucasians (E=1) to be male (G=1) than female (G=0) but more other ethnicities (E=0) to be female than male. On the other hand, we compose balanced validation and test sets since skewed validation/test sets give an advantage to specific biased models in evaluation.

\section{Appendix B: Comparison of standard accuracy and \textit{equalized~accuracy}}
\begin{table*}[h]
\caption{Comparison of standard accuracy (Acc.) and \textit{equalized~accuracy} (EAcc.). We assume a skewed test dataset and two biased classification models (Model A and Model B). Although the standard accuracy gives an advantage to a specific biased model (Model A) depending on the distribution of the test dataset, \textit{equalized~accuracy} measures a performance independent of the distribution.}
\label{table:appendixB}
\begin{center}
\resizebox{0.99\textwidth}{!}{
\begin{tabular}{ccccccccccccccc}
\toprule
\multicolumn{3}{c}{\textbf{Test Dataset}} &  & \multicolumn{5}{c}{\textbf{Model A}} &  & \multicolumn{5}{c}{\textbf{Model B}} \\
\multirow{2}{*}{TA} & \multicolumn{2}{c}{PA} &  & \multirow{2}{*}{TA} & \multicolumn{2}{c}{PA} & \multirow{2}{*}{Acc. $\uparrow$} & \multirow{2}{*}{EAcc. $\uparrow$} &  & \multirow{2}{*}{TA} & \multicolumn{2}{c}{PA} & \multirow{2}{*}{Acc. $\uparrow$} & \multirow{2}{*}{EAcc. $\uparrow$} \\ \cline{2-3} \cline{6-7} \cline{12-13}
 & P=1 & P=0 &  &  & P=1 & P=0 &  &  &  &  & P=1 & P=0 &  &  \\ \cline{1-3} \cline{5-9} \cline{11-15} 
T=1 & 4,000 & 1,000 &  & T=1 & 90\% & 10\% & \multirow{2}{*}{\textbf{74\%}} & \multirow{2}{*}{\textbf{50\%}} &  & T=1 & 10\% & 90\% & \multirow{2}{*}{\textbf{26\%}} & \multirow{2}{*}{\textbf{50\%}} \\
T=0 & 1,000 & 4,000 &  & T=0 & 10\% & 90\% &  &  &  & T=0 & 90\% & 10\% &  & \\
\bottomrule
\end{tabular}
}
\end{center}
\end{table*}
To demonstrate a contribution of \textit{equalized accuracy} more clearly, we assume a skewed test dataset and two extremely biased classification results of model A and B as shown in Table~\ref{table:appendixB}. In this circumstance, the standard accuracy is calculated as follows:
model A = $\frac{4}{10}\times90\% + \frac{1}{10}\times10\% + \frac{1}{10}\times10\% + \frac{4}{10}\times90\% = \textbf{74\%}$,~ model B =$\frac{4}{10}\times10\% + \frac{1}{10}\times90\% + \frac{1}{10}\times90\% + \frac{4}{10}\times10\% = \textbf{26\%}$. However, it is difficult to claim that model A performs better than model B even if the accuracy of model A is much higher. The result is due simply to the skewed test dataset and model A has the same accuracy with model B as 50\% on a fully balanced test dataset.  On the other hand, \textit{equalized accuracy} is formulated as follows: \text{$\frac{1}{4}[\text{TPR}_{p_0}+\text{TNR}_{p_0}+\text{TPR}_{p_1}+\text{TNR}_{p_1}]$} and calculated as follows: model A = $\frac{1}{4}[90\% +10\% +10\% +90\%]= \textbf{50\%}$,  model B = $\frac{1}{4}[10\% +90\% +90\% +10\%]= \textbf{50\%}$. Therefore, \textit{equalized accuracy} measures a performance independent of the distribution of a test dataset and has same effect as a balanced test dataset. 

\section{Appendix C: The detailed structures of networks}
\begin{table*}[h]
\caption{The structure of VAE.}
\label{table:appendixC_2}
\begin{center}
\resizebox{0.99\textwidth}{!}{
\begin{tabular}{lccccc}
\toprule
\multicolumn{1}{l|}{Network} & Name & Channel & Output size & Kernel size / Stride / Padding & Activation \\ \hline
\multicolumn{1}{l|}{\multirow{7}{*}{Encoder}} & Input & 3 & 64x64 & - & - \\ \cline{2-6} 
\multicolumn{1}{l|}{} & Conv1 & 32 & 32x32 & 4/ 2/ 1 & Relu \\ \cline{2-6} 
\multicolumn{1}{l|}{} & Conv2 & 32 & 16x16 & 4/ 2/ 1 & Relu \\ \cline{2-6} 
\multicolumn{1}{l|}{} & Conv3 & 64 & 8x8 & 4/ 2/ 1 & Relu \\ \cline{2-6} 
\multicolumn{1}{l|}{} & Conv4 & 64 & 4x4 & 4/ 2/ 1 & Relu \\ \cline{2-6} 
\multicolumn{1}{l|}{} & Conv5 & 256 & 1x1 & 4/ 1/ 0 & Relu \\ \cline{2-6} 
\multicolumn{1}{l|}{} & Conv6 & 120 & 1x1 & 1/ 1/ 0 & - \\ \hline
 &  &  &  &  &  \\ \hline
\multicolumn{1}{l|}{\multirow{6}{*}{Decoder}} & Conv7 & 256 & 1x1 & 1/ 1/ 0 & Relu \\ \cline{2-6} 
\multicolumn{1}{l|}{} & ConvTranspose1 & 64 & 4x4 & 4/ 1/ 0 & Relu \\ \cline{2-6} 
\multicolumn{1}{l|}{} & ConvTranspose2 & 64 & 8x8 & 4/ 2/ 1 & Relu \\ \cline{2-6} 
\multicolumn{1}{l|}{} & ConvTranspose3 & 32 & 16x16 & 4/ 2/ 1 & Relu \\ \cline{2-6} 
\multicolumn{1}{l|}{} & ConvTranspose4 & 32 & 32x32 & 4/ 2/ 1 & Relu \\ \cline{2-6} 
\multicolumn{1}{l|}{} & ConvTranspose5 & 3 & 64x64 & 4/ 2/ 1 & -
\\
\bottomrule
\end{tabular}
}
\end{center}
\end{table*}
\begin{table*}[h]
\caption{The structures of Discriminator/ Classifiers.}
\label{table:appendixC}
\begin{center}
\resizebox{0.80\textwidth}{!}{
\begin{tabular}{c|ccccc}
\toprule
Network & Name & Input size & Output size & Activation & Drop out \\ \hline
\multirow{6}{*}{\begin{tabular}[c]{@{}c@{}}Discriminator\\ / Classifiers\end{tabular}} & Fc1 & - & 60 & - & - \\ \cline{2-6} 
 & Fc2 & 60 & 1000 & Leaky relu & 0.2 \\ \cline{2-6} 
 & Fc3 & 1000 & 1000 & Leaky relu & 0.2 \\ \cline{2-6} 
 & Fc4 & 1000 & 1000 & Leaky relu & 0.2 \\ \cline{2-6} 
 & Fc5 & 1000 & 1000 & Leaky relu & 0.2 \\ \cline{2-6} 
 & Fc6 & 1000 & 2 & Leaky relu & 0.2\\ \bottomrule
\end{tabular}
}
\end{center}
\end{table*}
Table~\ref{table:appendixC_2} and Table~\ref{table:appendixC} show the detailed structures of VAE, discriminator, and classifiers. All models~\cite{vae,Higgins2017betaVAELB,pmlr-v80-kim18b,pmlr-v97-creager19a} used in our experiments have the same structures as above for fair evaluation.

\section{Appendix D: Visualization of disentangled representation}
\begin{figure*}[t]
  \centering
  \includegraphics[clip=true, width=0.99\textwidth]{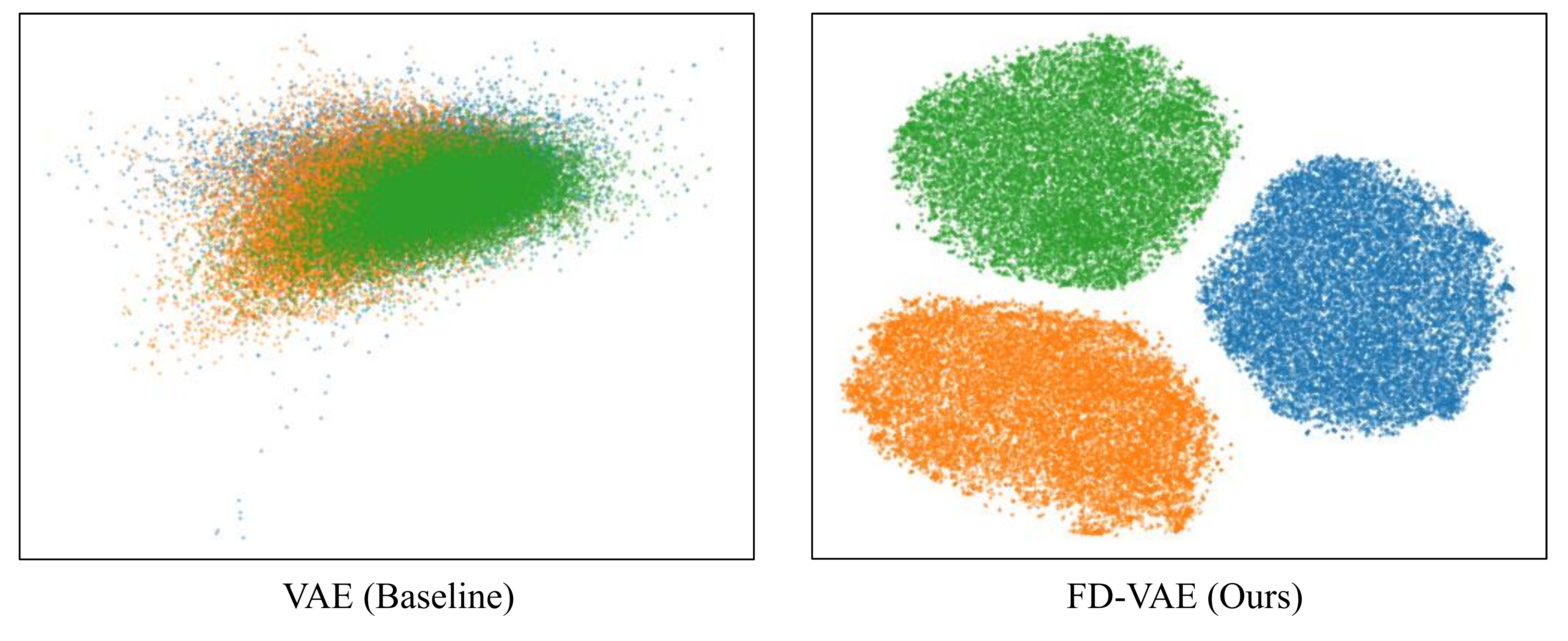}
  \caption{The t-SNE visualization on the subspaces of VAE (baseline) and FD-VAE (ours).The blue, green, and orange points denote TAL, PAL, and MAL in our model, respectively, and each divided space in VAE.}
  \label{fig:appendixD}
\end{figure*}
To validate that the subspaces of our model are effectively disentangled, we visualize them using t-SNE~\cite{t-sne}. Firstly, we learn VAE~\cite{vae} and FD-VAE on CelebA dataset~\cite{liu2015faceattributes}. In this process, we set the target and protected attributes to attractive and male attributes, respectively. Next, we use the learned models to get representation of all data in the test set. In addition, we separate the representation into three subspaces, which are used as data points in t-SNE (for VAE, we evenly divide it into three spaces that have the same dimensions with our subspaces). Figure~\ref{fig:appendixD} shows the t-SNE visualization on the subspaces. The blue, green, and orange points denote TAL, PAL, and MAL in our model, respectively, and each divided space in VAE. It shows that TAL, PAL, and MAL of our model are effectively disentangled.

\section{Appendix E: Computing infrastructure and implementation details}
We conduct all experiments on a single GTX 1080Ti GPU and develop all models~\cite{vae,Higgins2017betaVAELB,pmlr-v80-kim18b,pmlr-v97-creager19a} using PyTorch. In representation learning, the models are trained for about 120 epoch on CelebA dataset and 80 epoch on UTK Face dataset. The learning rate is set to $10^{-4}$ on both datasets. In downstream task learning, the models are trained for about 30 epoch and the learning rate is set to $10^{-6}$ on both datasets. The batch size is fixed at 256 in all steps.
{
\bibliography{neurips}
\bibliographystyle{abbrvnat}
}